\documentclass[conference]{IEEEtran}
\IEEEoverridecommandlockouts
\usepackage{cite}
\usepackage{amsmath,amssymb,amsfonts}
\usepackage{algorithmic}
\usepackage{graphicx}
\usepackage{subfig}
\usepackage{textcomp}
\usepackage{makecell}
\def\BibTeX{{\rm B\kern-.05em{\sc i\kern-.025em b}\kern-.08em
    T\kern-.1667em\lower.7ex\hbox{E}\kern-.125emX}}
\begin{document}

\title{Predicting Electric Vehicle Charging Station Usage: Using Machine Learning to Estimate Individual Station Statistics from Physical Configurations of Charging Station Networks
}

\author{\IEEEauthorblockN{Anshul Ramachandran}
\IEEEauthorblockA{\textit{Computer Science} \\
\textit{California Institute}\\
\textit{of Technology}\\
aramacha@caltech.edu}
\and
\IEEEauthorblockN{Ashwin Balakrishna}
\IEEEauthorblockA{\textit{Electrical Engineering} \\
\textit{California Institute}\\
\textit{of Technology}\\
abalakri@caltech.edu}
\and
\IEEEauthorblockN{Peter Kundzicz}
\IEEEauthorblockA{\textit{Computer Science} \\
\textit{California Institute}\\
\textit{of Technology}\\
pkundzic@caltech.edu}
\and
\IEEEauthorblockN{Anirudh Neti}
\IEEEauthorblockA{\textit{Electrical Engeineering} \\
\textit{California Institute}\\
\textit{of Technology}\\
aneti@caltech.edu}
}
% \and
% \IEEEauthorblockN{Ashwin Balakrishna}
% \IEEEauthorblockA{\textit{Electrical Engineering} \\
% \textit{California Institute of Technology}\\
% Pasadena, United States \\
% abalakri@caltech.edu}
% \and
% \IEEEauthorblockN{Peter Kundzicz}
% \IEEEauthorblockA{\textit{Computer Science} \\
% \textit{California Institute of Technology}\\
% Pasadena, United States \\
% pkundzic@caltech.edu}
% }

\maketitle

\begin{abstract}
Electric vehicles (EVs) have been gaining popularity due to their environmental friendliness and efficiency. EV charging station networks are scalable solutions for supporting increasing numbers of EVs within modern electric grid constraints, yet few tools exist to aid the physical configuration design of new networks. We use neural networks to predict individual charging station usage statistics from the station?s physical location within a network. We have shown this quickly gives accurate estimates of average usage statistics given a proposed configuration, without the need for running many computationally expensive simulations. The trained neural network can help EV charging network designers rapidly test various placements of charging stations under additional individual constraints in order to find an optimal configuration given their design objectives.
\end{abstract}


\section{Introduction}
Electric vehicles (EVs), particularly electric cars, promise to revolutionize the transportation sector due to no reliance on oil, low greenhouse emissions, and high energy efficiencies. As a result, by late 2016, 1 million EV units had been sold worldwide \cite{Shahan}. In addition, many countries have allocated research and development budgets, such as \$2.4 billion in the US, or provided subsidies, such as in China, to promote the advancement and adoption of EV technologies \cite{Obama} \cite{China}. 

All EV technologies depend on batteries that have limited capacities yet are rechargeable. Without radical breakthroughs in battery technology, options for recharging are to swap out batteries or reload from the electric network. Swapping out batteries is neither cost effective nor practical due to varying standards of batteries and EV regulators. Having personal individual electric vehicle supply equipment (EVSE) at homes or private workplace lots is not a scalable economic solution, nor is it technologically supportable without large-scale changes in modern electric grids. Therefore, EVSE networks, with multiple charging stations at the same geographical location, with a few at-location transformers distributing electricity from the main electric grid to the EVSEs, have emerged as promising, scalable solutions to the battery recharging problem.
In order for an EVSE network to service user needs in a timely manner, it is crucial to have algorithms that can efficiently distribute electricity within these EVSE networks under constraints of charging rates, maximum current draw from the main grid, etc. For this reason, many researchers have spent time analyzing and improving so-called scheduling algorithms, optimizing for different metrics such as EV wait time \cite{Qin} and distribution system load variance minimization \cite{Li}. These algorithms optimize the electric load distribution given a fully configured network. There has, however, been a lack of analysis in optimization of the physical network configuration itself, which is the direction that we choose to take in this work. 

Physical distribution of EVSEs within a network strongly affects individual EVSE loads and overall usages (e.g. EVSEs closer to lot entrances have higher usage than EVSEs farther away due to human parking behavior). However, there is no single ideal physical EVSE network design due to individual network constraints, such as number and position of transformers, number and location of entrances, and statistics of normal EV entrance and exit times, among many others. This being said, having estimates of EVSE usage given physical configuration of the network could help choose a final network design given these individual network constraints. In turn, this could additionally help optimize other technical features of the EVSE network, such as transformer-EVSE connections for load distribution and intelligent surge protection. 

However, it is computationally impractical to try every lot design in the combinatorial space of possible EVSE network physical configurations and repetitively apply scheduling algorithms to get estimates of individual EVSE usage, even if these algorithms have closed-form solutions. This is due to the computational complexity of these scheduling algorithms, which often depend on lengthy simulations and calculations that do not scale well with either number of EVs or EVSEs. 

Given the lack of any scalable closed-form solution, we propose a machine-learning based solution to this individual EVSE usage prediction problem, specifically using neural networks. A machine-learning solution has three main benefits. First is speed, as once a machine learning model is trained, its feed-forward evaluation is generally very fast, several orders of magnitude faster than running a full scheduling algorithm simulation. Second is generalization, in which we could train a network on a subset of possible physical EVSE network layouts and evaluate on the remaining with transferability of knowledge learned from the trained subset. Third is scalability with respect to the size of the EVSE network, since we can extract constant-size features of the physical location of each individual EVSE (such as local neighborhood and distance to closest entrance) to use as the input to the machine-learning model. In addition to these general benefits, in this EVSE usage prediction problem, we aim only for an estimate of individual EVSE usage, not exact values. Thus, while machine learning models do not give the exact load values as would be calculated by application of the scheduling algorithms, this drawback is unimportant.

Using real EV behavior data to generate EV arrival, departure, and desired electricity schedule patterns, algorithms to generate physical EVSE network layouts, and parking behavior simulators, we created simulated scheduling algorithm inputs. An Adaptive Charging Network (ACN) scheduling algorithm was then applied to this simulated data to generate training data. We then used a transformation of the physical EVSE network layout as the input to neural networks. These networks attempted to use this information to predict the individual EVSE usage values, which were known from the generated training data. We show both qualitatively and quantitatively that this trained model provides reasonable estimates of individual EVSE usage when evaluated on physical EVSE network layouts that the model was not trained on.

\section{Approaches and Methods}

\subsection{Pipeline}

Figure 1 contains the flowchart and description of the pipeline taken to generate training data of per-EVSE charging statistics (\textit{Network Simulation} module), train neural networks (\textit{Machine Learning} module), and use the trained models to display our results interactively (\textit{Application Interface} module). Algorithms used in individual steps of the \textit{Network Simulation} module can be found in Sections III.A to III.D. Description of input preprocessing methods and machine learning architectures can be found in Sections III.E and III.F respectively. 

\begin{figure*}[h]
		\centering
        {{\includegraphics[width=\textwidth]{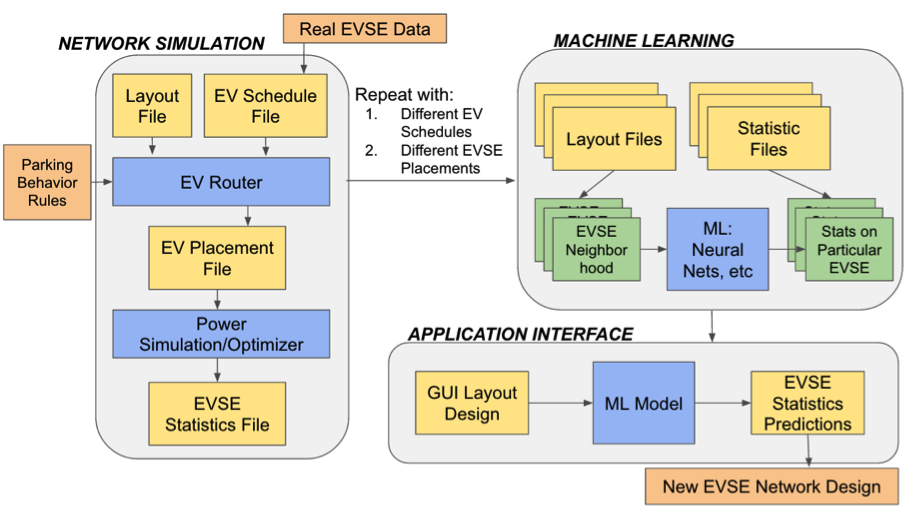} }}%
		\caption{\textbf{Flowchart of Steps Taken} All modules used to test the ability to predict individual EVSE statistics, such as power usage, based solely from the physical lot layout. The \textit{Network Simulation} module generates training data, the \textit{Machine Learning} module trains models to predict statistics from physical layout, and the \textit{Application Interface} module contains a user-facing portal to qualitatively visualize what the model predicts. In the \textit{Network Simulation} module, a routing algorithm uses a set of parking behavior rules to map EV and car arrivals, saved in the schedule file, to EVSEs and normal parking locations, saved in a provided physical lot layout file. With these assignments of EVs to EVSEs over time, the ACN linear programming solution calculates the statistics per EVSE. After accumulating these statistics using a large variety of incoming car schedules and physical layouts, the layouts are processed into input tensors for the neural network models in the \textit{Machine Learning} module, which predict, as output, the power usage statistic calculated from the ACN linear programming solution. In the \textit{Application Interface} module, a user can design a layout file using a Graphical User Interface (GUI), which then queries the trained model to display predicted statistics for the EVSEs in the designed layout. Testing multiple layouts can lead to a more informed EVSE physical layout configuration. Color Labels: Intermediate Files (Yellow), Overall System Input and Output (Orange), Algorithms and Models (Blue), Machine Learning Input and Output (Green).}
		\label{fig:flowchart}%
\end{figure*}

\subsection{Data Generation Parameters}

Using the \textit{Network Simulation} module, we generated a training set of 1215 size 30x30 layouts, each containing 15 EVSEs. For each layout, we generated and simulated 20 schedule files with 100 normal cars and 50 EVs. For each simulation, we calculate the average charging rate and overall power consumption over the 12-hour simulation period, and to calculate per-layout statistics, we average these metrics over the 20 schedules. We also generated a validation set of 141 size 30x30 layouts (i.e. models were not trained on any of these layouts), computing average metrics identically. On rare occasions, the garage layout generator algorithm resulted in a layout with no reachable EVSE. In these cases, the layout was regenerated.

\subsection{Machine Learning}

We chose to attempt to predict the average charging rate ($\tau$) and the total power consumption over a 12-hour period ($P_{tot}$) due to their importance in helping to balance loads. Neural networks were chosen since we have the ability to generate large amounts of simulated data and the input had many features to encode the full physical neighborhood of an EVSE in a layout. We chose Tensorflow as the backend for training due to optimized learning for neural network architectures, and used Keras, an open source package, as a wrapper around native Tensorflow. We used the mean squared loss as the loss function since both $\tau$ and $P_{tot}$ are continuous variables, an Adam optimizer, and a learning rate of 0.001.

\subsection{Web Application Portal}

The \textit{Application Interface} module allows for easy interaction with a trained neural network by allowing a user to design a layout within their web browser, evaluate the neighborhoods of EVSEs present through the trained model, and hover over the EVSE cells to view the predicted metrics. The web application was written in JavaScript and connects to an endpoint that interfaces with a trained model through Python Flask. Figure 2 shows a screenshot of the final portal.

\begin{figure}
		\centering
        {{\includegraphics[width=0.5\textwidth]{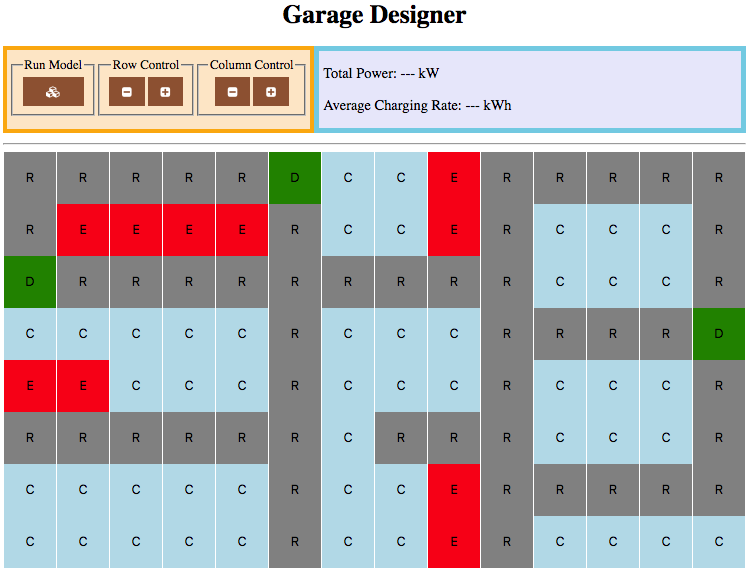} }}%
		\caption{\textbf{Screenshot of web application portal.} Users can click on individual cells to toggle the cell type to design the layout, run a trained model feed-forward, and intuitively display the individual EVSE statistics.}
		\label{fig:webshot}%
\end{figure}

\section{Algorithms and Models}

\subsection{Schedule Generation}

We obtained electric vehicle charging data from parking lots at Google Headquarters in Mountain View, CA and from parking garages at the California Institute of Technology. This data gives us multi-week data on arrival and departure times, energy demands, and peak charging rates per EV. Both the time parked and average charging rate (energy demanded/time parked) variables appeared to have an underlying normal distribution. To generate representative schedule files, we uniformly sample arrival times over a twelve-hour interval to correspond to EV arrivals. For each arrival, we first sample a peak charging rate from values that appeared in the real data. We also sample, using fitted normal distributions, a length of time parked (truncated to be nonnegative) and average charging rate (truncated to be nonnegative and at most the peak charging rate for the EV). From the length of time parked and average charging rate, we recalculate the energy demanded per EV. We save the arrival and departure times, energy demand, and peak charging rate for each EV. In addition, we similarly sample arrival and departure times  for normal cars, so the final schedule file contains a mixture of normal cars and EVs arriving and departing. \\

\subsection{Lot Physical Configuration Generation}

The only inputs for this module are the size of the garage (height and width, measured in cells) and $N$, the number of EVSEs desired in the charging network. The process of algorithmically generating physical layouts is demonstrated and explained in Figure 3 for the case when we desire a 5x5 layout containing $N=4$ EVSEs. \\

\begin{figure*}
		\centering
        {{\includegraphics[width=\textwidth]{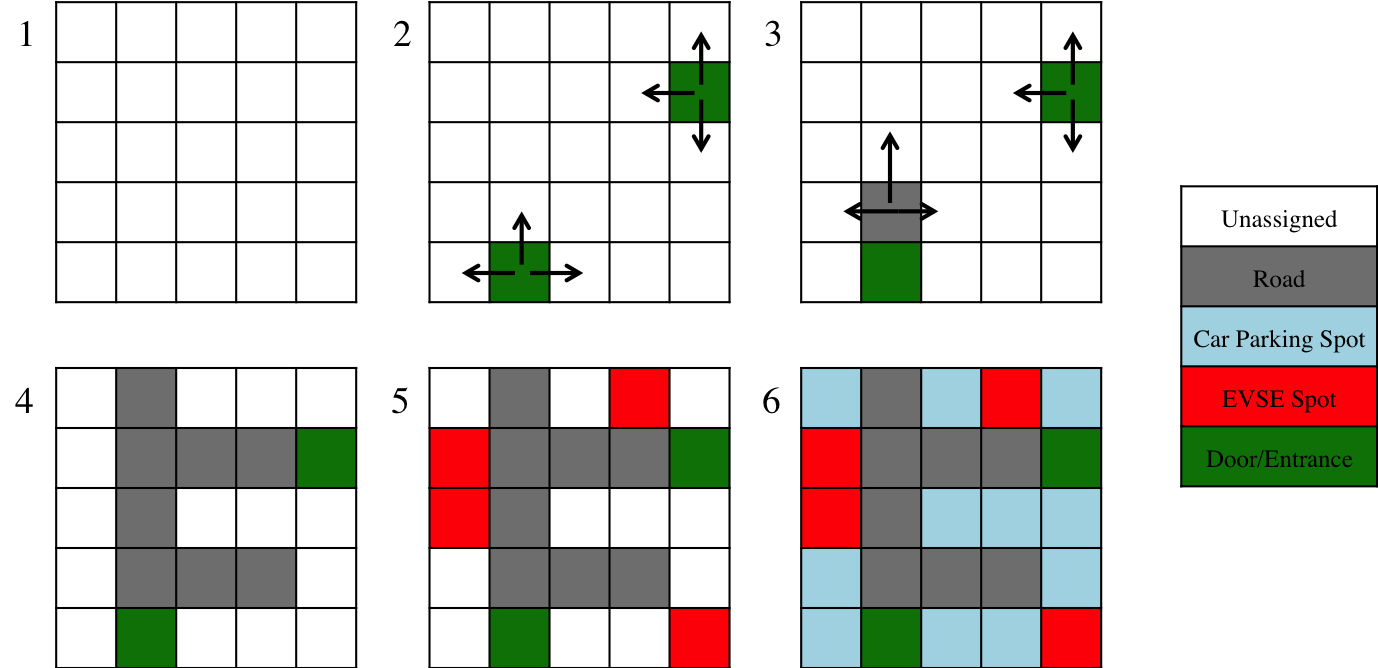} }}%
		\caption{\textbf{Steps in parking lot generation algorithm.} We start with a completely unassigned matrix of labels in the desired parking lot size (\#1). Cells on the edges of the matrix are independently chosen to be entrances to the parking lot with a fixed, small probability (\#2). We then ?grow? roads from each door, with each initial direction having equal probability (\#2). When an active road (road that has not halted) is extended, that road gets a high probability of continuing to extend in the same direction, with smaller probabilities for splitting (two new active roads form, one left and one right from the original direction) and halting (the road is no longer actively growing) (\#3). The longer a road gets, the higher probability of halting, and the more splits occur, the smaller the probability of splitting becomes. Halting also occurs when an active road attempts to extend past the edges of the matrix or into a door cell. All roads are finalized when no active roads exist (\#4). Out of the remaining unassigned cells, $N$ are uniformly randomly assigned to be EVSEs, with $N$ a provided input (\#5). The remaining unassigned cells are assigned as normal car parking spots (\#6).}
		\label{fig:lotgen}%
\end{figure*}

\subsection{Parking Simulator}

The parking simulator takes, as input, a lot physical layout file, a schedule file of incoming EVs and normal cars, and a set of parking behavior rules. In order of arrival in the schedule file, the parking simulator applies the following steps to determining the placement of the EV or normal car: 
\begin{enumerate}
  \item Random selection of which door the car/EV will enter from.
  \item Treat searching for empty parking spot as a depth-first search (DFS) in a road-tree network with root node at the door.
  \item Check if there are any free parking spots horizontally or vertically adjacent to the current door/road cell. This must match the type of car - normal parking spot for a normal car, EVSE for an EV.
  \item For each free parking spot, calculate probability of parking in the spot based off of the defined parking behavior rules. Park with calculated probabilities (independently if multiple free spots), and if parking occurs, this car has been successfully routed and the parking spot is marked as not free. Proceed to next car/EV arrival.
  \item If parking does not occur at the current door/road cell, proceed to the next door/road cell as would be in DFS. If there are multiple children (branching in the road network), then choose child with uniform probability (other branch will be explored if car is not parked in the chosen branch as in DFS).
  \item Repeat steps 3-5 until car/EV has been parked or until entire road network tree has been explored. If entire network has been explored without the car being parked (due to no valid free parking spots or not-park being probabilistically chosen repeatedly), skip the car/EV and proceed to the next (analogous to car/EV leaving the lot). 
\end{enumerate}

Note that we check the departure times for all currently parked cars and EVS between iterations of this process for each car/EV arrival. If any departure is scheduled to happen before the next arrival, the corresponding cell that that car/EV was parked in is remarked as free. The output of this parking simulator is an EV placement file, which is the mapping from EV in the schedule file to an EVSE in the layout file.

This approach for a parking simulator was chosen due to parallels to actual human behavior, as a person normally enters a lot and chooses a row through it, trying to find any empty spot along that row before perhaps retracing and trying a different row. The parking behavior rules used for collecting data for the models trained in this work includes smaller probabilities of parking in a free parking spot that has neighboring cells occupied and larger probabilities of parking in free parking spots near the edges of the lot.

Figure 4 demonstrates how our particular choice of parking behavior rules can affect the distribution of probabilities of a car/EV parking in particular locations.
\\

\begin{figure}
		\centering
        {{\includegraphics[width=0.4\textwidth]{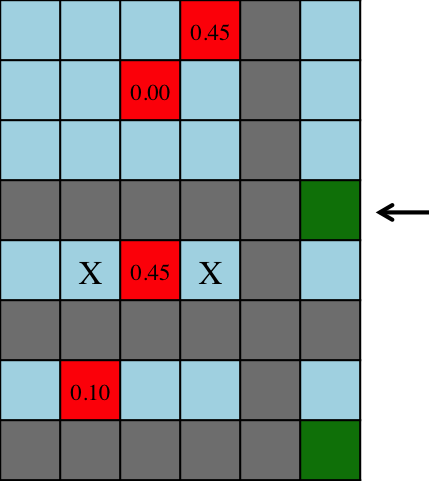} }}%
		\caption{\textbf{Parking simulator algorithm decision-making.} Probabilities shown are the overall probability of an EV incoming at the arrow-marked door parking at the particular EVSE conditional on the EV being parked instead of leaving the lot. Note that all EVSEs initially are free, but the normal parking spots marked with an X are occupied. We see that the EVSE closest to the entrance (in the center) has the same overall probability of being chosen by the depth-first search algorithm as the slightly farther EVSE at the top edge due to the former being surrounded by two occupied parking spots and latter being an edge spot next to an empty cell. The farther reachable EVSE at the bottom has a significantly smaller probability than either of these two EVSEs. The unreachable EVSE has a probability of zero of being chosen by the incoming EV, as expected.}
		\label{fig:parking}%
\end{figure}

\subsection{Power Allocation Scheduling Algorithm}

We use the Online Linear Program (OLP) version of the Adaptive Charging Network (ACN) solution for charging scheduling proposed by Lee, et al \cite{Lee} to calculate individual EVSE power usage statistics. We added modifications to explicitly map particular EVs to particular EVSEs at specific times along the simulation?s time horizon using the parking simulator?s output. We also introduced methods to extract power-related statistics from the network at the individual EVSE level. A small, example layout file and a corresponding charging rate profile timeline is shown in Figure 5 and a timeline for a more complex layout is shown in Figure 6 both with unconstrained max total network charging rate. \\

\begin{figure}
		\centering
        {{\includegraphics[width=0.4\textwidth]{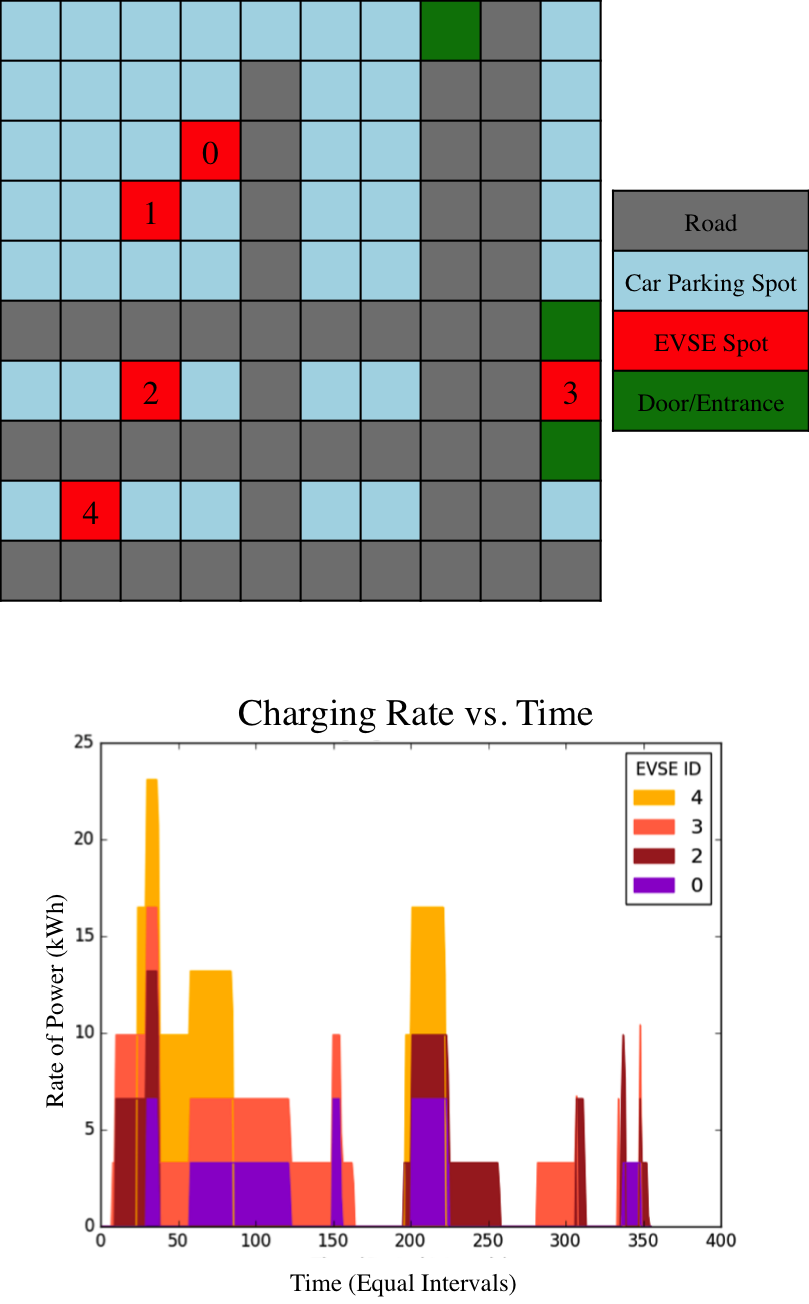} }}%
		\caption{\textbf{Small example parking lot configuration and associated charging rate profile graph.} As expected, EVSE \#3 near the doors is used for the longest time overall, and EVSE \#1 has zero charging as it is unreachable by the road/door network.}
		\label{fig:charging1}%
\end{figure}

\begin{figure}
		\centering
        {{\includegraphics[width=0.4\textwidth]{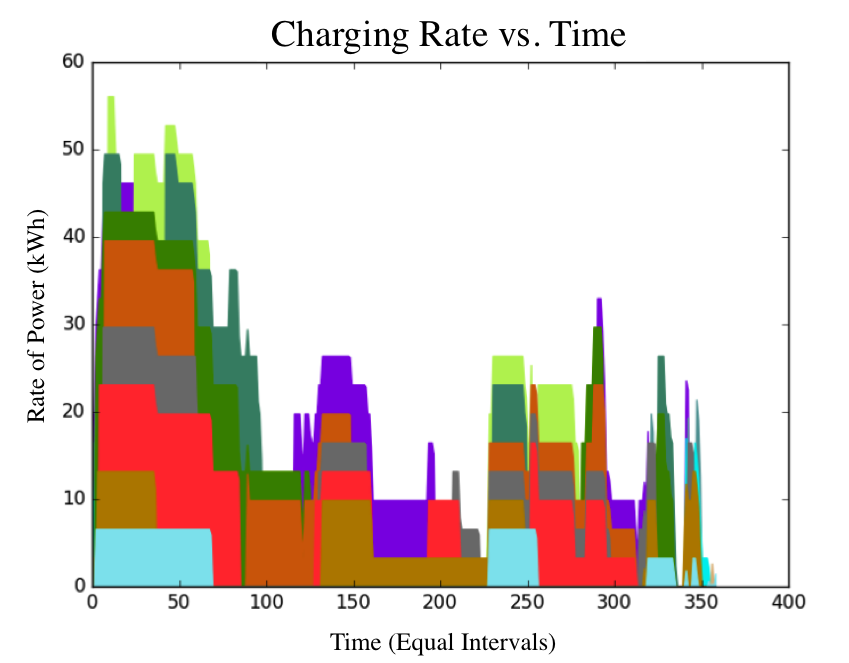} }}%
		\caption{\textbf{Example charging rate profile graph from a complex parking lot configuration.} 30x30 garage with 20 distinct EVSEs}
		\label{fig:charging2}%
\end{figure}

\subsection{Machine Learning Preprocessing}

Our goal was to predict an individual EVSE?s power consumption statistics based on its physical location and surroundings. Therefore, as input to our neural networks, we used the $M^2$ cells within the square with side length $M$ centered at the EVSE in question. We also expanded along the cell-type dimension to give us a 3D binary tensor of dimensions $(M, M, 5)$, with the 5 cell types being [road, normal parking spot, EVSE, door, off-grid]. Off-grid pertains to positions within the $M$x$M$ square that lie outside the edges of the matrix due to the EVSE being near the edge of the lot. We then flattened this 3D tensor into a 1D tensor of size $M * M * 5$. For Models 4 and 5, we also appended the minimum distance between the EVSE and any door as an additional term, due to our belief that this could be an important piece of information, especially for EVSEs that contain no door within its $M$x$M$ neighborhood. A simple example of this transformation with $M = 3$ is shown in Figure 7. For training models, we used $M = 9$. \\

\begin{figure*}
		\centering
        {{\includegraphics[width=\textwidth]{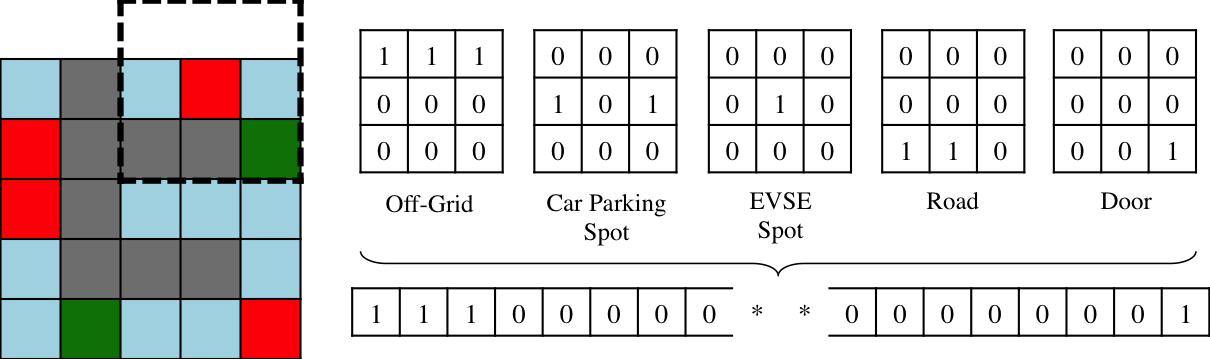} }}%
		\caption{\textbf{Example of converting the surrounding information for an EVSE cell to its input 1D tensor for the neural networks.} This is done by extracting the neighborhood of size $M = 3$ centered at the EVSE and identifying the types of cells within the neighborhood, generating a 3D tensor of shape $(M, M, 5)$ which is flattened into a 1D tensor, which is used as input.}
		\label{fig:charging2}%
\end{figure*}

\subsection{Machine Learning Architectures}

Figure 8 contains a table of the 5 combinations of neural network architecture, input, and output preprocessing tested. \\

\begin{figure*}
\centering
\begin{tabular}{c|c|c|c}
\textbf{Model \#} & \textbf{Architecture} & \textbf{Input} & \textbf{Output} \\\hline
1 & One hidden layer (128 nodes), fully connected  & \makecell{binary tensor,\\ size = $M * M * 5$} & Raw $\tau, P_{tot}$ \\ \hline
2 & One hidden layer (128 nodes), fully connected  & \makecell{binary tensor,\\ size = $M * M * 5$} & Log $\tau, P_{tot}$ \\ \hline
3 & One hidden layer (128 nodes), fully connected  & \makecell{binary tensor,\\ size = $M * M * 5$} & Log $\tau, P_{tot}$ \\ \hline
4 & One hidden layer (128 nodes), fully connected  & \makecell{binary tensor + $d_{door}$,\\ size = $M * M * 5 + 1$} & Log $\tau, P_{tot}$ \\ \hline
5 & Two hidden layers (256 nodes, 256 nodes), fully connected & \makecell{binary tensor + $d_{door}$,\\ size = $M * M * 5 + 1$} & Log $\tau, P_{tot}$ \\
\end{tabular}
\caption{\label{tab:variables} \textbf{Table of all 5 combinations of architectures parameters, input, and output tested.} Binary tensor refers to the output of the preprocessing of the layout file described in \textit{Machine Learning Preprocessing}. For all, the size of the neighborhood is $M = 9$. $d_{door}$ = distance to closest door, $\tau$ = average charging rate, $P_{tot}$ = total power consumed (within 12 hour period). }
\end{figure*}

\section{Results}
The mean squared error for each of the 5 combinations is shown in Figure 9. These plots were used to determine the best-tested model, chosen to be Model 3. We finally presented our Web Portal to Prof. Steven Low (California Institute of Technology), an expert in the field of EV networks and PI for the work done in developing the ACN charging solutions used in this work, to validate the predictions without knowledge of the underlying models, with positive reactions. Many qualitative and quantitative tests through the Web Portal, such as EVSEs closer to multiple doors having higher predicted overall power usage and EVSEs placed with no adjacency to the road/door network having zero predicted usage, were verified to be satisfied by the chosen underlying network.

\begin{figure*}
		\centering
   \subfloat[Model 1]
 {{\includegraphics[width=0.4\textwidth]{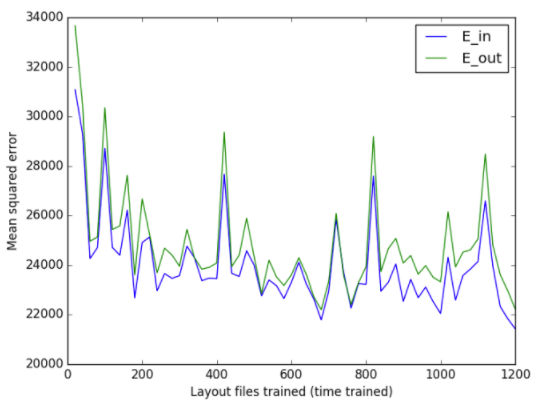} }}%

 \subfloat[Model 2]
 {{\includegraphics[width=0.4\textwidth]{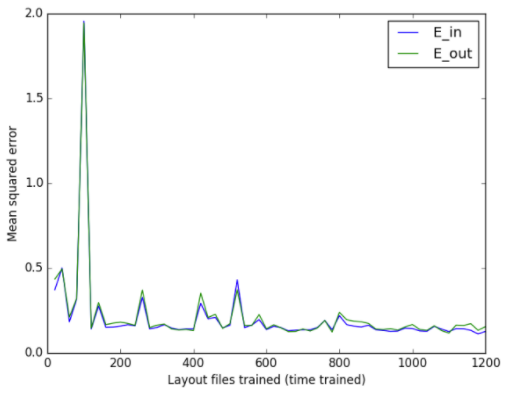} }}%
		\qquad
         \subfloat[Model 3]
 {{\includegraphics[width=0.4\textwidth]{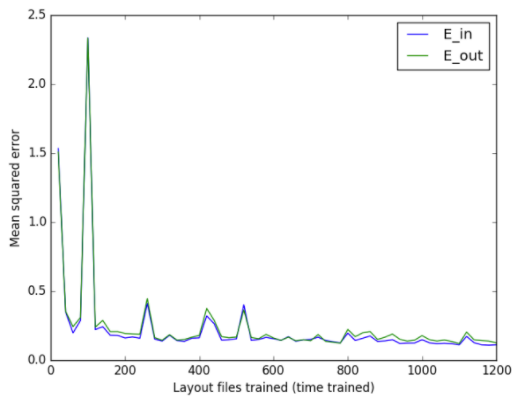} }}%
 
  \subfloat[Model 4]
  {{\includegraphics[width=0.4\textwidth]{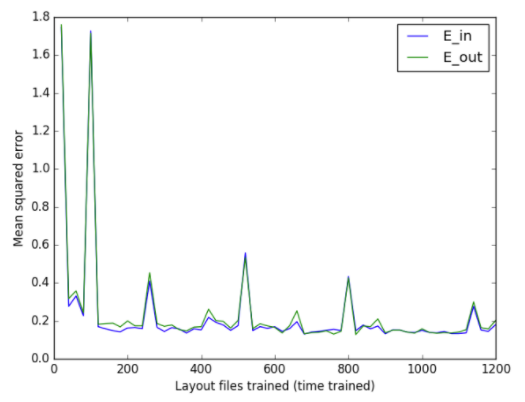} }}%
		\qquad
         \subfloat[Model 5]
 {{\includegraphics[width=0.4\textwidth]{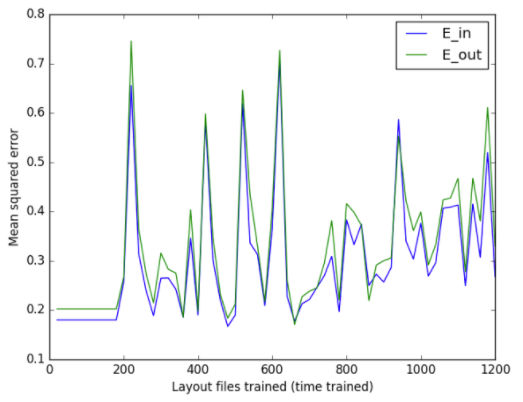} }}%
 
% 		\qquad
% {{\includegraphics[width=3cm]{lighting_images/4_shallownn} }}%
% 		\qquad
% {{\includegraphics[width=3cm]{lighting_images/4_shallowclassifier} }}%
%         \subfloat[Standard Model] {{\includegraphics[width=3cm]{lighting_images/0_base} }}%
% 		\qquad
% 		\subfloat[Narrow NN Base]{{\includegraphics[width=3cm]{lighting_images/0_narrow} }}%
% 		\qquad
% 		\subfloat[Shallow NN Base]{{\includegraphics[width=3cm]{lighting_images/0_shallownn} }}%
% 		\qquad
% 		\subfloat[Shallow Classifier]{{\includegraphics[width=3cm]{lighting_images/0_shallowclassifier} }}%
		 
		\caption{\textbf{Trial Training Curves}  In sample and out of sample mean squared error over number of epochs trained for total power usage prediction for each of the 5 combinations of neural network architecture, input, and output preprocessing.}
		\label{fig:training}%
\end{figure*}

\section{Discussion and Conclusion}
The error plots demonstrated that none of the models showed signs of overfitting, with the out of sample error from evaluating the model on layouts in the validation dataset never straying too far above the in sample error throughout the training process. Discussions of each model are as follows:
\begin{enumerate}
\item Directly predicting raw values, with target values ranging over many multiple orders of magnitude, likely led to large, fluctuating gradient steps and lack of convergence.
\item Predicting the natural logarithm of raw target values led to convergence, likely due to more stable gradient steps. Mean squared error of logarithm values is equivalent to square of logarithm of the raw predicted value divided by the target predicted value, and therefore errors around 0.2-0.3 mark improvement in predicting raw values as compared to directly doing so in Model 1. 
\item Additional hidden units showed a small improvement in out of sample error without causing overfitting.
\item Adding a distance to nearest door input increased the in and out of sample errors, contrary to initial belief. It is possible that the distance term was on a different order of magnitude as the remaining binary values or simply further training would be required to reach lower errors.
\item A second hidden layer likely created far too many tunable weights in the fully connected architecture for the scale of data used to train. The low absolute values of error combined with lack of convergence signify that memorization to specific patterns in the generated layouts occurred. This would likely not generalize if different layout generation algorithms were also used in the data generation pipeline.
\end{enumerate}

Model 3 performed best in terms of a combination of low absolute error, convergence, and lack of over-fitting. Therefore, it was chosen as the backend to the Web Portal for our quantitative and qualitative tests and demonstration to Prof. Low. The positive results in the portal based tests and positive reaction from Prof. Low suggests that there is merit to this approach towards predicting power statistics for individual EVSEs solely from the layout, and while the overall problem was simplified in this work, the approach could be extended and become an important design tool.

There are many possible improvements, extensions, and complexities that could be added to each step in the Network Simulation module.
\begin{itemize}
  \item \textit{Schedule Generation}: using real data to sample inter-arrival car times, finding max peak charging rates by contacting the manufacturers of various EVs, vary numbers of EVs and cars that arrive in total over a 12 hour period
  \item \textit{Layout Generation}: additional types of tiles (one way roads, stairs/exits, etc) for realism, multiple levels in the layout as in a parking garage, restrictions on distance between EVSE and transformer
  \item \textit{Parking Simulator}: additional rules of parking behavior such as minimizing distance to stairs or repeated use of a particular EVSE by a single user over multiple days, additional types of behavior (e.g. traveling in circles rather than DFS)
  \item \textit{Charging Simulator}: collect data under different charging algorithms, train alternate model for each charging algorithm used \\
\end{itemize}

For machine learning, large improvements can be achieved solely from collecting more training data by varying parameters used in the schedule and layout generation steps. In terms of neural network architecture, we would definitely test more combinations of architecture and input preprocessing. In addition, layout matrices share parallels to images, so incorporating convolutional layers, which have proven powerful in helping to extract information from images, is an intriguing future exploration. We could also explore alternate classes of models other than neural networks. Finally, we would like to quantitatively validate on real life data, which would require digitally encoding the layouts of already constructed lots.

Using this work on predicting power usage metrics given layout, we could also create tools for helping search the possible layout space under user-set constraints to find optimal layouts. Overall, this work has uncovered the power of applying machine learning in problems related to EV charging networks, which show promise to dramatically change our electric and transportation infrastructures.

\end{document}